\documentclass[sigconf,nonacm]{acmart}

\usepackage{booktabs}
\usepackage{longtable}
\usepackage{amsmath}

\usepackage{amssymb}
\usepackage{enumitem}
\usepackage{graphicx}
\graphicspath{{./}}
\usepackage{xcolor}
\usepackage{tikz}
\usetikzlibrary{arrows.meta}
\definecolor{lrn}{RGB}{200,110,0}
\definecolor{frz}{RGB}{30,80,160}

\setcopyright{none}
\renewcommand\footnotetextcopyrightpermission[1]{}
\newcommand{\eOnestformed}{3}

\newcommand{\eOnesoftfour}{0.02--0.30}

\newcommand{\eOnebcursofttwentyfour}{0.97--1.00}
\newcommand{\eOnebcursoftformed}{4}

\newcommand{\eOnebmixsoftformed}{2}
\newcommand{\eOnebcurstsix}{0.01--0.12}

\newcommand{\eOnebcursoftTwoxtwentyfour}{0.92--1.00}

\newcommand{\xTwobmformed}{2}

\newcommand{\eTwodirectFourSix}{0.35--0.38}

\newcommand{\eTwodirectOneTwo}{0.10--0.12}
\newcommand{\eTwoeTwoeFourSix}{0.37}

\newcommand{\eTwosupFourSix}{1.00}
\newcommand{\eTwosupFourSixheld}{0.55--0.59}
\newcommand{\eTwosupOneTwo}{1.00}
\newcommand{\eTwoetwoeexact}{0.00}
\newcommand{\eTwosupswap}{1.00}
\newcommand{\eTwosupreaderoff}{0.07--0.10}
\newcommand{\eTwobscratch}{0.34--0.41}
\newcommand{\eTwobwarmstexact}{0.00}
\newcommand{\eTwobwarmauxexact}{0.00--0.28}

\newcommand{\eTwobwarmrl}{1.00}

\newcommand{\xOneprior}{0.95--1.00}
\newcommand{\xOnepriorexact}{0.30--0.54}
\newcommand{\xOnepriorswap}{0.38--0.60}

\newcommand{\xOnefrozenlow}{0.25--0.28}

\newcommand{\xOnebpassed}{1}
\newcommand{\xOnebruns}{4}
\newcommand{\xOnebstuck}{0.67--0.71}
\newcommand{\xOneckOneSixpassed}{3}
\newcommand{\xOnecmempassed}{2}
\newcommand{\xOneckOneSixmempassed}{2}

\newcommand{\xOnectwelve}{1.00}

\newcommand{\xFourtFourkEightheld}{0.79--0.86}

\newcommand{\xFourtOneTwokEightheld}{0.96--0.97}

\newcommand{\xFourtFourkOneThreerange}{0.85--0.92}

\newcommand{\xFourkthirteenexact}{0.08--0.38}
\newcommand{\xTwoaonedirect}{0.13}

\newcommand{\xTwoafourcot}{0.93}

\newcommand{\xTwobdirectbbhseven}{0.78--0.83}

\newcommand{\xTwobsupexact}{0.00}

\newcommand{\xTwocall}{1.00}
\newcommand{\xTwocreaderoff}{0.15--0.24}
\newcommand{\xTwocrlten}{0.54--0.62}

\newcommand{\xThreelayer}{20}
\newcommand{\xThreetool}{235}
\newcommand{\xThreebig}{3091}
\newcommand{\xThreeratio}{11.6}
\newcommand{\xThreebigratio}{152}
\newcommand{\xThreebigacc}{0.94}

\newcommand{\xThreetoolheld}{0.93--0.98}

\newcommand{\xThreebkernelratio}{11.6}
\newcommand{\xThreebkernelratiosixteen}{8.6}

\newcommand{\xThreeblayerextra}{1.4}
\newcommand{\prOnemodule}{64}
\newcommand{\prOnewriterreader}{0.61}

\newcommand{\prOnegemma}{1.0}

\newcommand{\pThreewarmreinmaxexact}{0.00}

\newcommand{\pThreewarmimleexact}{0.00}

\newcommand{\pThreewarmall}{0.17--0.29}
\newcommand{\pThreescratch}{0.24--0.30}

\newcommand{\eTwosupheldplease}{1.00}

\newcommand{\eTwosupheldafter}{0.09}
\newcommand{\eTwosupheldafterexact}{0.00}
\newcommand{\xTwocseeds}{3}

\newcommand{\xThreeclayerkernel}{11.0}

\newcommand{\xThreeclayerkernelthree}{10.2}
\newcommand{\xThreectoolaccseven}{0.99}

\newcommand{\xThreectooltokens}{17}

\newcommand{\xThreectoolratio}{10}

\newcommand{\xThreectoolratiob}{3}

\newcommand{\xThreecOnebpotaccseven}{0.05}

\newcommand{\xThreecFourbcotaccseven}{0.93}

\newcommand{\xThreecFourbcottokens}{534}

\newcommand{\xThreecFourbcotratio}{621}

\newcommand{\xThreecFourbcotratiob}{186}

\newcommand{\xThreecFourbpotaccseven}{0.76}

\newcommand{\xThreecFourbpottokens}{252}

\newcommand{\xThreecFourbpotratio}{345}

\newcommand{\xFivearithFourSixalone}{1.00}

\newcommand{\xFivearithFourSixheldalone}{0.39--0.73}

\newcommand{\xFiveswapbbhsevenalone}{1.00}

\newcommand{\xFivereaderoffarith}{0.07}
\newcommand{\xFivereaderoffswap}{0.14}
\newcommand{\xFivemmlubase}{0.382}
\newcommand{\xFivemmlu}{0.368--0.370}

\newcommand{\xFivefiremmluarith}{0.050--0.090}
\newcommand{\xFivefiregeneralarith}{0.072--0.090}
\newcommand{\xFivefiremmluswap}{0.042--0.058}

\newcommand{\xFivebmmlu}{0.376--0.386}

\newcommand{\xFivebxfivevalidarithmmlu}{3.4--5.4\,\%}
\newcommand{\xFivebxfivevalidswapmmlu}{2.8--4.0\,\%}

\newcommand{\xFivebqneganyarithmmlu}{6.6--11.4\,\%}
\newcommand{\xFivebqneganyswapmmlu}{0.4--0.8\,\%}

\newcommand{\xFivebqnegvalidlogic}{36.0--38.0\,\%}
\newcommand{\xFivebqnegvalidgsm}{28.2--37.2\,\%}

\newcommand{\xFiveblone}{88 of 91}

\newcommand{\xFivecstrictlogic}{0.0--4.8\,\%}
\newcommand{\xFivecstrictgsm}{0.2--0.6\,\%}

\newcommand{\xFivecgatemmlu}{0.382}
\newcommand{\xFivecgatemmlufresh}{0.392}

\newcommand{\xFivecprompts}{3{,}800}
\newcommand{\xFivecbbh}{0.99--1.00}
\newcommand{\xSixpassed}{5}
\newcommand{\xSixruns}{6}
\newcommand{\xSixbbh}{1.00}
\newcommand{\xSixreaderoff}{0.14}

\newcommand{\xSixstep}{250--1000}
\newcommand{\xSixgamebbh}{0.00--0.20}
\newcommand{\xSixgamesyn}{1.00}
\newcommand{\xSevenkept}{234}
\newcommand{\xSevenrewrites}{300}
\newcommand{\xSevenparser}{0}
\newcommand{\xSevenlayerlorathree}{0.43--0.81}

\newcommand{\xSevenlayerloraseven}{0.27--0.85}

\newcommand{\xSevenlayersupdrop}{0.12--0.13}

\newcommand{\xSevenlayerfreeseven}{0.90--0.94}

\newcommand{\xSevenlayerfreedrop}{0.06--0.10}

\newcommand{\xSevenlayergatedfire}{0.39--0.53}

\newcommand{\xSevenlayergatedwherefired}{0.93--1.00}

\newcommand{\xSeventoolseven}{0.68}
\newcommand{\xSeventoolsevenorig}{0.99}

\newcommand{\xSevencotFourbseven}{0.71}
\newcommand{\xSevencotFourbsevenorig}{0.96}

\newcommand{\xSevencotFourbdrop}{0.26}
\newcommand{\xSevennoadapters}{0.87--0.96}

\newcommand{\xSevenbpos}{400}
\newcommand{\xSevenbneg}{400}

\newcommand{\xSevenbgbothfire}{73.4--88.3\,\%}
\newcommand{\xSevenbgbothacc}{0.82--0.86}

\newcommand{\pSixpassed}{2}
\newcommand{\pSixruns}{4}
\newcommand{\pSixreaderoff}{0.08}
\newcommand{\pSixmmlufire}{0.0--1.2\,\%}

\newcommand{\pSixbasemmlu}{0.576}
\newcommand{\pSixmmlu}{0.576--0.582}
\newcommand{\pSixbbh}{1.00}

\newcommand{\pSixopen}{0.90--0.95}
\newcommand{\pSixopenseven}{0.90--0.95}

\newcommand{\pSixfailbbh}{0.17--0.95}
\newcommand{\pSixkernel}{24.7}
\newcommand{\pSixplain}{23.0}
\newcommand{\pSixoverhead}{1.07}
\newcommand{\pSixcotratio}{278}

\newcommand{\pSixcotwallratio}{299}
\newcommand{\xSevencgatepass}{87.2--93.6\,\%}
\newcommand{\xSevencwherefired}{0.94--1.00}
\newcommand{\xSevenccalsets}{7}
\newcommand{\xSevencothermax}{-4.7}
\newcommand{\xSevencgbotht}{0.68--0.74}
\newcommand{\xSevencgbothfire}{65.8--81.8\,\%}
\newcommand{\xSevencgbothacc}{0.78--0.84}
\newcommand{\xSevencgbothnear}{0.0\,\%}

\begin{document}

\title{A Tool in the Forward Pass: Frozen Tape Modules as Layers of a Small Language Model}

\author{Michael Granitzer}
\affiliation{\institution{University of Passau}\city{Passau}\country{Germany}}
\author{Saber Zerhoudi}
\affiliation{\institution{University of Passau}\city{Passau}\country{Germany}}
\author{Mohammad Al-Maamari}
\affiliation{\institution{University of Passau}\city{Passau}\country{Germany}}
\author{Hai Dang Dang}
\affiliation{\institution{University of Passau}\city{Passau}\country{Germany}}
\author{Kanishka Ghosh Dastidar}
\affiliation{\institution{IT:U Interdisciplinary Transformation University Austria}\city{Linz}\country{Austria}}
\renewcommand{\shortauthors}{Granitzer et al.}

\begin{abstract}
Calling a tool through generated text adds decoding steps, even when the computation itself is short. We study an
alternative: a small neural module that runs inside a language model's forward pass. A learned writer turns hidden
states into a \emph{symbolic tape}, a short sequence of discrete symbols that encodes the computation. The module,
trained separately and then frozen, reduces the tape to a result, and a learned reader returns that result to the
model. We evaluate arithmetic and object tracking in gemma-3-1b-it. With a writer trained on labelled tapes, the layer
answers all 750 items of BIG-Bench Hard's tracking shuffled objects correctly. At seven people it needs about
$1/\xThreectoolratio$ of the GPU kernel time of a text-tool baseline that uses the same host model and exact code,
with accuracy \xTwocall\ against \xThreectoolaccseven. The writer can also be learned from answers alone, guided by a
hand-written copying prior, but less reliably: \xSixpassed\ of \xSixruns\ runs succeed in the 1B host and
\pSixpassed\ of \pSixruns\ in the 4B host. Two independently trained layers keep their task accuracy when attached to
one model, and a learned gate keeps them silent on the tested general-text and look-alike prompts. The results support
a narrow use case: a bounded computation that returns one symbol can avoid the decoding cost of a textual tool call.
Extracting the right input reliably, and deciding when to act, remain the main open problems.

\end{abstract}

\begin{teaserfigure}
\centering
\begin{tikzpicture}[font=\sffamily\scriptsize,
  ttl/.style={anchor=north west, font=\sffamily\small\bfseries},
  sub/.style={anchor=north west, text=black!65, font=\sffamily\tiny},
  bx/.style={draw=#1!70, fill=#1!7, rounded corners=2pt, align=center, inner sep=2.5pt},
  code/.style={anchor=north west, draw=black!30, fill=black!3, rounded corners=1pt, inner sep=3pt,
    font=\ttfamily\tiny, align=left, text width=#1},
  cell/.style={draw=black!45, fill=white, minimum width=0.3cm, minimum height=0.34cm, inner sep=0, font=\sffamily\scriptsize},
  arr/.style={-{Stealth[length=1.4mm]}, thick, black!55}]
\node[anchor=north west, draw=black!35, fill=black!2, rounded corners=2pt, text width=17.3cm, inner sep=4pt, align=left]
  at (0,0) {\textbf{One item} (tracking shuffled objects, BIG-Bench Hard): \emph{Alice, Bob and Claire hold a red, a blue
  and a green ball. First, Alice and Bob swap balls. Then, Claire and Alice swap balls. Finally, Bob and Claire swap
  balls. At the end of the game, Bob has the}\ \ (A) red ball\ \ (B) blue ball\ \ (C) green ball.\ \ Answer: (B).};
\begin{scope}[yshift=-1.1cm]
\node[ttl] at (0,0) {(a) A tool as a program};
\node[sub] at (0,-0.38) {program of thought: the model writes code, an interpreter runs it};
\node[code=5.3cm] (pa) at (0,-0.75) {\textcolor{lrn}{balls = \{"Alice": "red", "Bob": "blue",\\
\ \ \ \ \ \ \ \ \ "Claire": "green"\}\\
for a, b in [("Alice", "Bob"), ("Claire", "Alice"),\\
\ \ \ \ \ \ \ \ \ \ \ \ \ ("Bob", "Claire")]:\\
\ \ \ \ balls[a], balls[b] = balls[b], balls[a]\\
print(balls["Bob"])}};
\node[bx=frz, anchor=north west] (ia) at (0,-2.55) {Python interpreter};
\node[anchor=west] at (2.35,-2.72) {$\rightarrow$ ``blue'' $\rightarrow$ (B)};
\node[anchor=north west, text width=5.5cm, font=\sffamily\tiny] at (0,-3.15)
  {Every generated token is one forward pass: \xThreecFourbpottokens\ tokens per item at seven people
   (gemma-3-4b-it, accuracy \xThreecFourbpotaccseven).};
\end{scope}
\begin{scope}[xshift=6.0cm, yshift=-1.1cm]
\node[ttl] at (0,0) {(b) A tool through text (baseline)};
\node[sub] at (0,-0.38) {the model writes the tool's input; hand-written code computes};
\node[code=5.3cm] (pb) at (0,-0.75) {\dots\ At the end of the game, Bob has the\\
Call:\textcolor{lrn}{\ Alice\ Bob\ Claire\ Alice\ Bob\ Claire\ Bob\ =}};
\foreach \i in {0,...,7} \node[draw=lrn!70, fill=lrn!15, minimum width=0.42cm, minimum height=0.16cm, inner sep=0]
  at (0.75+0.6*\i,-1.75) {};
\node[anchor=west, font=\sffamily\tiny, text=black!65] at (0,-2.0) {8 generated tokens = 8 more forward passes};
\node[bx=frz, anchor=north west] (cb) at (0,-2.55) {swap code (hand-written)};
\node[anchor=west] at (2.75,-2.72) {$\rightarrow$ Bob $\rightarrow$ (B)};
\node[anchor=north west, text width=5.5cm, font=\sffamily\tiny] at (0,-3.15)
  {Learned by adapters on our own stories: \xThreectooltokens\ tokens per item at seven people (gemma-3-1b-it,
   accuracy \xThreectoolaccseven).};
\end{scope}
\begin{scope}[xshift=12.0cm, yshift=-1.1cm]
\node[ttl] at (0,0) {(c) A tool as a layer (this paper)};
\node[sub] at (0,-0.38) {inside the one forward pass that reads the prompt};
\node[bx=black, minimum width=1.25cm] (lo) at (0.62,-1.0) {layers 1--8};
\node[bx=lrn] (wr) at (2.05,-1.0) {writer};
\draw[arr] (lo) -- (wr);
\foreach \s [count=\i] in {A,B,C,A,B,C,B} \node[cell] (c\i) at (2.65+0.3*\i,-1.0) {\s};
\draw[arr] (wr) -- (c1);
\node[sub, anchor=west] at (4.85,-0.88) {tape};
\node[bx=frz] (md) at (3.85,-1.75) {module (frozen)};
\draw[arr] (c4.south) -- (md.north);
\node[bx=lrn] (rd) at (2.05,-1.75) {reader};
\draw[arr] (md) -- (rd);
\node[circle, draw=black!55, inner sep=0.5pt] (pl) at (0.62,-1.75) {$+$};
\draw[arr] (rd) -- (pl); \draw[arr] (lo) -- (pl);
\node[bx=black, minimum width=1.25cm] (up) at (0.62,-2.45) {layers 9--26};
\draw[arr] (pl) -- (up);
\node[anchor=west] at (0.05,-3.0) {$\rightarrow$ (B)};
\node[anchor=north west, font=\ttfamily\tiny, align=left, text=black!75] at (2.3,-2.1)
  {round 1: A B C A [B C B] $\rightarrow$ C\\ round 2: A B [C A C] $\rightarrow$ A\\ round 3: [A B A] $\rightarrow$ B
   \ \ (Bob's own)};
\node[anchor=north west, text width=5.5cm, font=\sffamily\tiny] at (0,-3.15)
  {Writer and reader learned, module trained alone; gemma frozen. One forward pass
   (gemma-3-1b-it, accuracy \xTwocall).};
\end{scope}
\begin{scope}[yshift=-5.2cm]
\node[ttl] at (0,0) {(d) The tape layer in detail};
\node[anchor=north west, inner sep=0] at (0,-0.45) {\includegraphics[width=11.2cm]{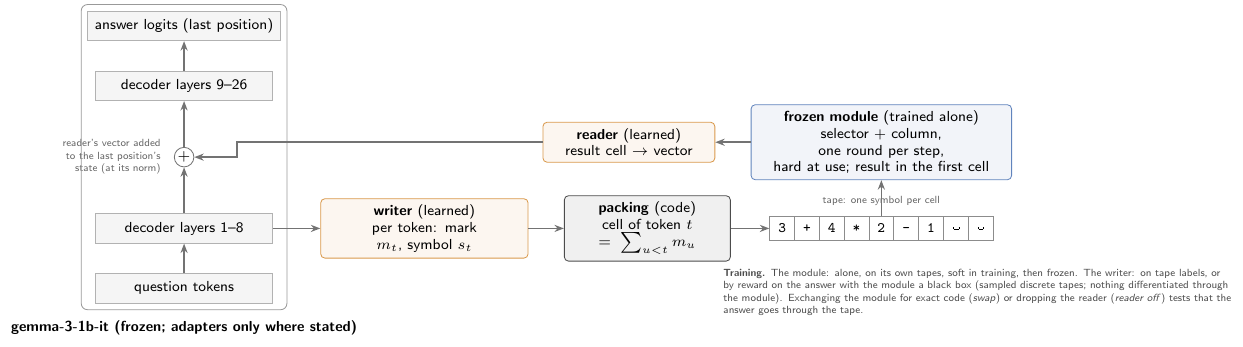}};
\end{scope}
\begin{scope}[xshift=11.6cm, yshift=-5.2cm]
\node[ttl] at (0,0) {(e) Result: near-equal accuracy, less cost};
\node[sub, text width=5.9cm] at (0,-0.38) {accuracy shown at seven people; GPU time per item at seven people relative
  to the layer (log scale)};
\foreach \n/\r/\a/\c [count=\i] in {{tape layer, 1B}/1/\xTwocall/lrn,
    {text tool, 1B}/\xThreectoolratio/\xThreectoolaccseven/black,
    {program of thought, 4B}/\xThreecFourbpotratio/\xThreecFourbpotaccseven/black,
    {chain of thought, 4B}/\xThreecFourbcotratio/\xThreecFourbcotaccseven/black} {
  \pgfmathsetmacro\L{0.08+0.85*log10(\r)}
  \node[anchor=east, align=right, font=\sffamily\tiny] at (2.55,-0.75-0.5*\i) {\n\\[-1pt] accuracy \a};
  \fill[\c!55] (2.65,-0.86-0.5*\i) rectangle ++(\L,0.24);
  \node[anchor=west, font=\sffamily\tiny] at (2.68+\L,-0.75-0.5*\i) {\r$\times$};}
\node[anchor=north west, text width=5.9cm, font=\sffamily\tiny] at (0,-3.05)
  {Separate experiments: answer-only training succeeds in some runs; two gated layers share one frozen model
   and preserve its answers on the tested negative prompts.};
\end{scope}
\end{tikzpicture}
\caption{\textbf{One computation, three interfaces.} (a) The model generates a program for an interpreter.
(b) It generates only the inputs to a fixed tool. (c) A learned writer constructs those inputs directly from hidden
states. A frozen neural module computes the result, and a reader returns it to the model. The tape traces Bob's
final object back through the swaps to its original owner. (d) The layer's components. (e) Measured GPU kernel time
at seven people, relative to the layer, with each method's accuracy (\S\ref{sec:eff}).}
\label{fig:overview}
\end{teaserfigure}

\maketitle

\section{Introduction}

A language model that uses a calculator, an interpreter or a database writes the call as text, waits for the result and
reads it back~\cite{schick2023toolformer,gao2023pal,chen2023pot}. This interface is flexible, but generating the call requires sequential decoding steps. The returned result must
also be processed by the model, although its tokens can be processed together. For a short computation inside a longer
answer --- arithmetic in a word problem, the state of a few objects in a story --- the call can cost more than the answer.

A tool can instead sit inside the forward pass (Figure~\ref{fig:overview}, a--b against c). A writer reads hidden states and hands the
tool its input, the tool computes, and a reader writes the result back, all in one pass. Earlier work has shown this
with exact code as the tool and a writer supervised on annotated operands~\cite{dietz2025igc,dhanraj2025neurosymbolic},
and with a symbolic network steered from hidden states~\cite{dugan2024occamllm}. We examine four questions in this setting. Can the tool be
a \emph{learned} module, trained on its own and frozen, so that modules can be developed and exchanged independently
(RQ1, and RQ5 for several modules)? What does a tool in the forward pass save against the same tool used through text,
at near-equal accuracy (RQ2)? How can the model learn what to hand the tool when nobody labels the tool's input (RQ3)?
And does the interface read wording it was never trained on (RQ4)?

We study these questions with a \emph{tape layer} (Figure~\ref{fig:overview}c, d). The intuition is a scratchpad that
the model fills in without writing a word: instead of spelling a computation out as text, the model places its inputs
on a short row of cells inside the forward pass, a small specialised network computes on that row, and the result is
handed back to the model's hidden state.

Conceptually, the row is a \emph{symbolic tape}: a short sequence of cells, each holding one discrete symbol from a
small alphabet, with a \emph{tape language} that fixes which tapes are well formed and what they mean. A \emph{tape
module} is a small network that rewrites the tape step by step, each step replacing a few neighbouring cells by one,
until a single cell holds the result. The module is trained alone, on tapes only, and then frozen. The \emph{tape
layer} connects it to a language model: a learned writer reads the hidden states and fills the tape, the frozen module
computes, and a learned reader adds the result back to the hidden state from which the model answers.

Concretely, we use two tasks, each with its own tape language and module, in gemma-3-1b-it~\cite{gemma3}. In
\emph{arithmetic}, the tape is the expression itself, such as \texttt{3+4*2}, and the module evaluates it. This is the
development task on which the layer was designed. In \emph{tracking shuffled objects} from BIG-Bench Hard (BBH), a benchmark
of hard reasoning tasks~\cite{suzgun2023bbh}, people exchange objects and the question asks what one of them holds at
the
end. The tape lists who swaps with whom, followed by the person asked about (Figure~\ref{fig:overview}c). We chose this task after developing the arithmetic architecture. Its benchmark items are excluded from training,
but later interface decisions were informed by experiments on this task. It is therefore a transfer case study, not
a completely untouched test of the final design.

The central result is a cost reduction at near-equal accuracy: the layer performs the computation during prompt
processing, while the text-tool baseline generates its inputs token by token. Our contributions are:
\begin{itemize}[leftmargin=*]
\item \textbf{A learned tool that the model uses (RQ1).} The tool is a small network trained alone on tapes, then
  frozen. One check shows that it computes correctly on the tapes the model writes: replacing it by exact code of the
  same computation gives the same accuracy. A second shows that the answer depends on it: removing its result from
  the hidden state drops accuracy to chance (\S\ref{sec:layer}).
\item \textbf{Length extrapolation (RQ1).} The module applies one small rule round after round, so a longer input
  needs more rounds, not new parameters. Trained on up to six operators, it reaches \eOnebcursofttwentyfour\ at
  twenty-four. As a layer, the model answers \eTwosupOneTwo\ at twelve operators, where the same model with adapters
  alone answers \eTwodirectOneTwo, and it is exact on seven exchanges after training on at most five
  (\S\ref{sec:layer}).
\item \textbf{A tool call without generated tokens (RQ2).} On seven-person object tracking, the layer is as accurate
  as the same model calling exact code through text (\xTwocall\ against \xThreectoolaccseven) at
  $1/\xThreectoolratio$ of its GPU kernel time per item at batch size one (\S\ref{sec:eff}).
\item \textbf{Which tokens to hand the tool, learned from answers (RQ3).} Without labelled tapes, rewarding tapes
  whose result is right trains the writer: \xSixpassed\ of \xSixruns\ runs in gemma-3-1b-it and \pSixpassed\ of
  \pSixruns\ in gemma-3-4b-it become exact. A hand-written prior supplies each token's symbol (digits and operators, a
  fixed name-to-index lexicon), so reward learns which tokens to mark, for example to leave out the story's
  introduction, not what the symbols mean. Gradients through the frozen module destroy the writer instead
  (\S\ref{sec:writer}).
\item \textbf{Unseen wording (RQ4).} On benchmark items rewritten by another language model, writers on the frozen
  model reach \xSevennoadapters, while gemma-3-4b-it with chain of thought falls from \xSevencotFourbsevenorig\ to
  \xSevencotFourbseven\ at seven people (\S\ref{sec:open}).
\item \textbf{Tools added independently (RQ5).} Two layers trained separately, each for its own task, share one
  frozen model. Each stays as accurate as alone, and with a learned gate neither acts on any of \xFivecprompts\ general
  and look-alike prompts, so accuracy on MMLU, a multiple-choice knowledge benchmark~\cite{hendrycks2021mmlu}, equals
  the base model's (\S\ref{sec:two}).
\end{itemize}
The claim concerns two task-specific tape languages, fixed computation budgets and single-symbol outputs. It does
not establish a general tool interface or an advantage of learned execution over exact code. On both tasks exact
code would have served as well. The learned module is a feasibility result for tools that have no code, or that
should be trained and exchanged independently. Registered predictions and their outcomes, including failures, are
listed in Appendix~\ref{app:pred}.

\section{Related work}

\textbf{Tools as text.} Toolformer learns API calls written into the text~\cite{schick2023toolformer}. Program-aided
models generate code for an interpreter~\cite{gao2023pal,chen2023pot}. With GPT-4, this approach solves the
object-tracking benchmark~\cite{zhang2024nlep}. These methods are accurate, and their cost is the generated tokens.

\textbf{Tools in the forward pass.} IGC places a gated calculator after layer 1 of a frozen Llama, with a writer
trained on annotated operands~\cite{dietz2025igc}. Dhanraj and Eliasmith encode a hidden state into symbolic vectors by
supervised regression, compute in code and decode the result back~\cite{dhanraj2025neurosymbolic}. OccamLLM steers a
symbolic network from hidden states, trained by REINFORCE~\cite{dugan2024occamllm}. CALM joins a frozen model to a
second frozen model by learned cross-attention, and TransNAR joins one to a frozen graph
reasoner~\cite{bansal2024calm,bounsi2024transnar}. Neither writes a discrete input for the attached module. In
contrast, our tool is a learned module, the writer emits a tape of variable length, and we measure the cost against the
same tool used through text.

\textbf{Programs from answers.} Learning which program to hand an executor from the answer alone is the setting of
neural symbolic machines and their successors~\cite{liang2017nsm,liang2018mapo,guu2017mml,yi2018nsvqa}. Their finding
--- reward learning needs a warm start or a memory of rewarded programs, and spurious programs win otherwise ---
reappears at our hidden-state interface. Gradients through discrete steps are biased~\cite{bengio2013ste}. Estimators
with lower bias exist~\cite{niepert2021imle,liu2023reinmax}, and we test two of them. Separately trained neural
subroutines composed by fixed control flow are the precedent for our module~\cite{yan2020nee}, and neural module
networks compose separately defined modules into a network per question~\cite{andreas2016nmn}.

\textbf{Reasoning without tokens.} Implicit and latent reasoning also avoid generating intermediate tokens, by
distilling a chain of thought into the hidden states~\cite{deng2023implicitcot} or by feeding the last hidden state
back as the next input~\cite{hao2024coconut}. There the model itself computes. In our setting a separate module does,
and its input is a discrete tape that can be inspected and checked.

\section{The tape layer}
\label{sec:method}

\textbf{Intuition.} The module computes as one would simplify an arithmetic expression by hand. It chooses a small
part of the tape, replaces that part with its result, and repeats. For example, it reduces \texttt{3+4*2} first to
\texttt{3+8}, then to \texttt{1} under arithmetic modulo ten. The language model's job is to identify the expression
in the prompt and read back the result. The module's job is to carry out the computation.

\textbf{Tape updates.} Let $T^{(r)}$ be the tape after round $r$, a sequence of $N$ cells over a task-specific
alphabet $\Sigma$. In each round a selector $S_\theta$ chooses an inner cell or the no-operation action $\bot$:
\begin{equation}
i_r = \arg\max_{i \in \{2,\dots,N-1\} \cup \{\bot\}} S_\theta\bigl(T^{(r)}\bigr)_i .
\end{equation}
If $i_r=\bot$, the tape stays as it is. Otherwise a column $C_\theta$ computes a replacement symbol from the chosen
cell and its two neighbours, and a hand-written splice replaces those three cells by it and closes the gap:
\begin{align}
z_r &= \arg\max_{z\in\Sigma} C_\theta\bigl(T^{(r)}_{i_r-1},\,T^{(r)}_{i_r},\,T^{(r)}_{i_r+1}\bigr)_z,\\
T^{(r+1)} &= \operatorname{splice}\bigl(T^{(r)},i_r,z_r\bigr).
\end{align}
The module runs a fixed number of rounds $K$, so rounds after the result has been reached must choose $\bot$.

\textbf{Network and read scope.} In this study the column is a multilayer perceptron, not a transformer. It
concatenates the embeddings of three adjacent cells, applies two hidden linear layers with GELU activations, and
predicts a result symbol. The selector uses a one-dimensional convolution over seven cells (radius three), a GELU
and a pointwise output layer. A separate network scores the no-operation action from the mean tape embedding.
Three cells is the smallest window that holds a binary operation written on a tape: two operands and an operator.
Unary and binary operations suffice in principle, because an operation with more arguments can be built from them
with more rounds and intermediate symbols. In the state-tracking language the column applies a three-argument rule
directly (a swap pair and the queried person), since it fits the same window. The restriction that matters is
therefore not the number of arguments but locality: the operands must stand next to each other on the tape, and the
tape language must place them there. The selector's wider window supplies neighbouring context for choosing a reduction. We do not establish
that radius three is minimal or optimal. One column is reused at every location and round within each task-specific
module. Thus its input is a triple of cells, not a single cell, but it cannot directly select two arbitrary distant
cells. General nonlocal addressing and a library of different columns are outside this experiment.

\textbf{Module training.} The module has \prOnemodule\,k parameters. It is trained from input tapes and final result
tapes, without labels for intermediate choices. During training, distributions over symbols and selections replace
the discrete decisions above (the \emph{soft} module). A curriculum gradually increases tape length. At evaluation, every selection and
replacement uses an argmax: the module is \emph{used hard}. The module is then frozen before training its interface to the language model.

\textbf{The two tape languages.} For arithmetic (expressions mod 10 with $+$, $-$ and $\times$) the tape is the
expression, a round reduces one operator with its two operands, and the module must learn operator precedence. For
state tracking the tape is the swapped pairs' person indices followed by the person asked about, and a round reduces
the rightmost triple $(a, b, x)$ to $b$ if $x=a$, to $a$ if $x=b$, and to $x$ otherwise: it follows the asked
person's object backwards through the swaps. The remaining cell is the person who first held what the asked person ends
with (Figure~\ref{fig:overview}c).

\textbf{From prompt to tape and back} (Figure~\ref{fig:overview}d). The frozen module sits after decoder layer 8 of
the 26 of gemma-3-1b-it, about a third of its depth. The position trades two needs. The writer needs states that
already tell which tokens are names, digits or operators. The upper layers need enough depth left to turn the
injected result into an answer. We tested one alternative: a writer reading layer 13 writes fewer exact tapes and is
less accurate in range (Appendix~\ref{app:more}). In gemma-3-4b-it the layer sits at the same relative depth (layer
10 of 34). The position was not tuned further and is a parameter for follow-up work. For every prompt token $t$, the writer, a small learned network on the hidden state $h_t$,
gives a mark $m_t\in\{0,1\}$ and a symbol $s_t\in\Sigma$. In training, $m_t$ is a probability and $s_t$ a
distribution. A hand-written cumulative sum gives each marked token the cell $j_t=\sum_{u<t}m_u$, so the initial
tape holds the marked tokens' symbols in prompt order:
\begin{equation}
T^{(0)}_{j_t}=s_t \quad\text{for every } t \text{ with } m_t=1,
\end{equation}
and blanks in the remaining cells. The writer must therefore also leave out tokens that look like tape symbols but do
not belong to the computation. Our prompts contain few such distractors. One arithmetic training template names an
unrelated day number, and each story names all its people in the introduction before any exchange. Prompts with many
unrelated numbers or names are left to later work. Let $F_\theta$ denote the module's $K$ rounds and $R_\psi$ the learned reader,
which maps the result cell to a vector scaled to the norm of the hidden state. The layer updates the hidden state at
the last prompt position $n$,
\begin{equation}
h'_n=h_n+g\,R_\psi\bigl(F_\theta(T^{(0)})\bigr),
\end{equation}
and the upper layers answer from $h'_n$ with a digit or an option letter. With a single layer, $g=1$. In the
composition experiments, $g=1$ only if a learned prompt gate accepts the prompt and the tape passes the
tape-language and capacity checks. Each layer decides separately, and if neither acts, the host is unchanged. The two
layers never call each other. Where stated, LoRA adapters~\cite{hu2022lora} on every layer let the model adapt to the
result. Without them, gemma is used exactly as released, and only the writer and reader (\prOnewriterreader\,M
parameters, against \prOnegemma\,B frozen) are trained.

\textbf{Training the reader.} The reader is trained by the answer loss only: the cross-entropy of the right answer
token, back-propagated through the frozen upper layers. Each symbol's reader vector starts as gemma's own embedding of
that symbol's token, so the upper layers can read the result from the first step.

\textbf{Training the writer.} We generate the training prompts ourselves, so the true tape of each one is known: the
expression, or the exchanged names followed by the asked name. These are the \emph{tape labels}. With them, the writer
is trained by cross-entropy against the true tape, beside the answer loss. Without them, we sample $k$ discrete tapes per question, run each through the frozen module, and
reward those whose result is the right answer. The writer is updated by REINFORCE~\cite{williams1992reinforce} or by
maximum marginal likelihood over the rewarded samples~\cite{guu2017mml}. Nothing is differentiated through the module.
Three hand-written aids are optional. A \emph{prior} fits the writer, for the first 100 steps, to mark tokens that look
like tape symbols and copy them: digits and operators for arithmetic, a fixed name-to-index lexicon for state
tracking. A \emph{validity check} rewards only well-formed tapes. A \emph{memory} keeps each question's last rewarded
tape and trains its likelihood~\cite{liang2018mapo}.

\section{Experimental design}
\label{sec:design}

\textbf{Research questions.} Each question has its own subsection in \S\ref{sec:results}.
\begin{enumerate}[label=\textbf{RQ\arabic*},leftmargin=*,noitemsep]
\item Can a module trained alone be frozen and used as a layer, with the answer going through its tape?
\item What does the layer cost against the same tool used through text, at near-equal accuracy?
\item Can the writer be learned from answers alone, and what does that need?
\item Does the layer read wording it was never trained on?
\item Can layers trained separately share one model without interfering with it or with each other?
\end{enumerate}

\textbf{Tasks.} The \emph{development task} is arithmetic: expressions mod 10 inside sentence templates, trained on one
to six operators, tested in range and at twelve. The \emph{held-out task} is tracking shuffled objects from BIG-Bench
Hard~\cite{suzgun2023bbh}: three, five or seven people exchange objects in turn, and the question asks what one of them
holds at the end (750 test items). The benchmark items are excluded from training: layers are trained on synthetic stories with four scenarios and
one to five swaps. The task was chosen after the arithmetic architecture, but subsequent changes to writer
supervision, training and gating used findings from this task. We use ``held-out'' below only for the benchmark
items and wording, not for an untouched task throughout development. gemma-3-1b-it (1B below) answers it at chance
without reasoning (\xTwoaonedirect\ at seven people, chance 0.14), so the task has room for a tool.

\textbf{Baselines.} The \emph{adapters-only} baseline uses the same host and adapters without a tape. It tests
whether the training task can be solved without the tool. \emph{Text tool}: the same model, taught by adapters to write the tool's input as text after the
question, with the exact code computing the answer --- the closest comparison at near-equal accuracy. The call contains only the inputs, without function names, argument names or JSON syntax, which keeps it as short
as possible. Other tool formats and serving systems were not measured. \emph{Chain of thought} and
\emph{program of thought}~\cite{chen2023pot} with gemma-3-1b-it and gemma-3-4b-it (4B below), zero-shot.

\textbf{Checks that the answer goes through the tape.} Accuracy alone cannot show this, because the model may answer
without the tool. Each check answers one part. \emph{Tapes exact}: the share of written tapes identical to the true
tape, so the writer hands over the right input. \emph{Swap}: accuracy when the learned module is replaced by exact
code, so the module computes correctly on the tapes written. \emph{Reader off}: accuracy when the result is not added
back. It should fall to chance, so the answer depends on the module.

\textbf{Protocol.} gemma-3-1b-it runs in bf16 on one RTX 4090. Layers are trained for 2500 steps at batch 16, keeping the
checkpoint with the best validation accuracy. Two seeds per configuration, three for the main rows and four per module.
Each split has 300--400 test questions. Time is measured as wall-clock with the standard decoder and as GPU kernel time from
the profiler, which an overhead-free decoder could at best approach. Code, configurations with the registered
predictions, all results and the frozen modules are published on Zenodo~\cite{tapelayer2026zenodo}.

\section{Results}
\label{sec:results}

\subsection{RQ1: A frozen module supplies the answer}
\label{sec:layer}

\textbf{The module alone.} Trained only on its final tape, the arithmetic module learns the order of operations
without being told which operator comes first (Table~\ref{tab:eone}): with the length curriculum \eOnebcursoftformed\ of
4 seeds are exact at six operators, and they hold at twenty-four (\eOnebcursofttwentyfour), four times the trained
length. The same architecture and training loop, unchanged, learn state tracking (\xTwobmformed\ of 4 seeds, exact to
fourteen swaps).

\begin{table*}[t]
\caption{\textbf{Modules trained alone, from their final tape only} (four seeds each, used hard). Accuracy at the
longest trained length and at four times it (sixteen digits, twenty-four operators; $^{*}$fourteen swaps, 2.3 times).
``Formed'': seeds exact at the longest trained length. Hand-written: the tape layout, the splice and the number of
rounds.}
\label{tab:eone}
\small
\begin{tabular}{@{}llcccc@{}}
\toprule
module, trained from its final tape only & trained on & at the longest trained & at 4$\times$ & 16-place carry chain & formed \\
\midrule
addition, discrete tape (straight-through) & 1--4 digits & 0.08--1.00 & 0.00--1.00 & 0.00--1.00 & 3 of 4 \\
\quad + a continuous state per cell & 1--4 digits & 1.00 & 0.98--1.00 & 0.99--1.00 & 4 of 4 \\
\quad trained soft, used hard & 1--4 digits & 0.02--0.30 & 0.00 & -- & 0 of 4 \\
multiplication (either tape) & 1--4 digits & 0.00 & 0.00 & -- & 0 of 8 \\
operator order: selector + column, length curriculum & 1--6 operators & 1.00 & 0.97--1.00 & -- & 4 of 4 \\
\quad no curriculum & 1--6 operators & 0.19--1.00 & 0.12--1.00 & -- & 2 of 4 \\
\quad straight-through & 1--6 operators & 0.01--0.12 & 0.00--0.09 & -- & 0 of 4 \\
\quad twice the rounds at use & 1--6 operators & 1.00 & 0.92--1.00 & -- & 4 of 4 \\
state tracking (swaps): the same recipe, unchanged & 1--6 swaps & 0.92--1.00 & 0.80--1.00$^{*}$ & -- & 2 of 4 \\
\bottomrule
\end{tabular}
\end{table*}

\textbf{The module as a layer.} With tape labels for the writer, the layer answers \eTwosupFourSix\ at four to six
operators and \eTwosupOneTwo\ at twelve, twice the trained length, where the same model with adapters alone answers
\eTwodirectFourSix\ and \eTwodirectOneTwo\ (Table~\ref{tab:etwo}).
Replacing the learned module by exact code leaves
accuracy at \eTwosupswap, and switching the reader off drops it to \eTwosupreaderoff. On the held-out task the layer
is likewise exact on every split (\xTwocall\ for accuracy, tapes and swap, all \xTwocseeds\ seeds), also at seven
exchanges where training stopped at five, and with the reader off it falls to chance (\xTwocreaderoff;
Table~\ref{tab:state}).

\textbf{Extrapolation.} Both results above lie beyond the training lengths: twelve operators is twice the longest
training expression, and seven exchanges exceed the five of the training stories. The adapters alone do not
extrapolate (\eTwodirectOneTwo\ at twelve operators). The layer does because the module learned one local rule,
applied once per round, and a longer tape only needs more rounds. The language model has to write the longer tape, but
not to compute on it.

\textbf{What we learn.} A module trained alone can be frozen inside a model and carry the computation that the
model answers from. Exchanging it for exact code leaves accuracy unchanged, so the module computes correctly on the
tapes written, and without the reader accuracy falls to chance, so the answer depends on it. These checks are
necessary, not decorative. On the held-out task, the writer at
first marked nearly every token of the prompt and filled the tape with the story's introduction. No tape was exact
(\xTwobsupexact), yet accuracy was high, because the adapters answered alone: they reach \xTwobdirectbbhseven\ at
seven people without any tape. Only after the writer was taught which tokens to mark (the names in the exchanges and
in the question) did the answers go through the tape. This repair was decided on the held-out task.

\begin{table*}[t]
\caption{\textbf{The frozen module as a layer of gemma-3-1b-it on the development task, by how its writer is trained}
(two seeds; three for the first row, the tape-labels row and the last label-free row; used hard; trained on one to six
operators). ``Swap'': accuracy with exact code in place of the learned module; ``reader off'': with the result not added
back. Unseen templates: two sentence templates never trained on; $^{\dagger}$a set of eight, six of them new, not
comparable with the rows above; $^{\ddagger}$adapters below the writer frozen. \textit{(code)}: hand-written.}
\label{tab:etwo}
\small
\resizebox{\textwidth}{!}{\begin{tabular}{@{}llccccccc@{}}
\toprule
how the writer is trained & tape labels & 4--6 ops & unseen templates & 12 ops & tapes exact & swap & reader off \\
\midrule
adapters only, no tape & -- & 0.35--0.38 & 0.28--0.34 & 0.10--0.12 & -- & -- & -- \\
\midrule
answers only, gradient through the soft module & none & 0.37 & 0.29--0.31 & 0.10--0.12 & 0.00 & 0.28--0.36 & 0.12 \\
answers only, straight-through & none & 0.34--0.41 & 0.29--0.33 & 0.09 & 0.00 & 0.32--0.34 & 0.11--0.16 \\
answers only, reward (module a black box) & none & 0.36--0.39 & 0.30--0.31 & 0.07--0.11 & 0.00 & 0.32--0.34 & 0.14--0.17 \\
answers only, ReinMax$^{\ddagger}$ & none & 0.24--0.30 & 0.17--0.18 & 0.12 & 0.00 & 0.20--0.27 & 0.20--0.25 \\
answers only, I-MLE$^{\ddagger}$ & none & 0.24--0.25 & 0.16--0.20 & 0.08--0.10 & 0.00 & 0.20--0.21 & 0.18--0.21 \\
600 labelled steps, then straight-through & 600 steps & 0.28--0.32 & 0.25--0.28 & 0.12 & 0.00 & 0.25--0.35 & 0.11--0.13 \\
\quad the same, lower adapters frozen$^{\ddagger}$ & 600 steps & 0.23--0.29 & 0.18 & 0.09--0.13 & 0.00 & 0.29 & 0.20--0.28 \\
600 labelled steps, then ReinMax$^{\ddagger}$ & 600 steps & 0.19--0.25 & 0.14--0.18 & 0.10 & 0.00 & 0.17--0.18 & 0.16--0.26 \\
600 labelled steps, then I-MLE$^{\ddagger}$ & 600 steps & 0.17--0.20 & 0.14--0.16 & 0.06--0.08 & 0.00 & 0.11--0.17 & 0.17--0.18 \\
600 labelled steps, then answers + 0.1 tape loss & 600 steps & 0.33--0.47 & 0.26--0.35 & 0.12--0.17 & 0.00--0.28 & 0.33--0.47 & 0.11 \\
100 labelled steps, then reward & 100 steps & 1.00 & 0.55 & 1.00 & 1.00 & 1.00 & 0.08--0.11 \\
tape labels throughout & all & 1.00 & 0.55--0.59 & 1.00 & 1.00 & 1.00 & 0.07--0.10 \\
\midrule
copy prior \textit{(code)}, then reward & none & 0.95--1.00 & 0.14--1.00 & 0.79--1.00 & 0.30--0.54 & 0.38--0.60 & 0.08--0.11 \\
\quad reward only for well-formed tapes \textit{(code)} & none & 0.67--1.00 & 0.33--0.55 & 0.36--1.00 & 0.30--1.00 & 0.60--1.00 & 0.12--0.17 \\
\quad + 16 sampled tapes per question & none & 1.00 & 0.55--0.75 & 1.00 & 1.00 & 1.00 & 0.07--0.10 \\
\midrule
tape labels throughout, 12 training templates$^{\dagger}$ & all & 1.00 & 0.96--0.97 & 1.00 & 1.00 & 1.00 & 0.08--0.10 \\
\quad 4 templates, writer after layer 13$^{\dagger}$ & all & 0.85--0.92 & 0.84--0.86 & 0.86--0.93 & 0.11--0.29 & 0.85--0.92 & 0.10--0.12 \\
\bottomrule
\end{tabular}}
\end{table*}

\subsection{RQ2: The cost at near-equal accuracy}
\label{sec:eff}

The closest baseline uses the same host model to generate the inputs for exact code. Its accuracy is close to the
layer's, so the comparison is about cost. On arithmetic the layer answers in \xThreelayer\,ms per question, the text tool in \xThreetool\,ms
(\xThreeratio$\times$), and gemma-3-4b-it with chain of thought in \xThreebig\,ms (\xThreebigratio$\times$) at lower
accuracy (\xThreebigacc; Table~\ref{tab:eff}). Measured in GPU kernel time, which removes launch overhead, the ratio is
the same (\xThreebkernelratio$\times$ at batch 1, \xThreebkernelratiosixteen$\times$ at batch 16).

On the held-out task the tools must write more, and the gap widens (Table~\ref{tab:bbheff}). The text tool achieves near-equal accuracy (\xThreectoolaccseven\ at seven people) after \xThreectooltokens\ generated tokens and
needs \xThreectoolratio$\times$ the layer's GPU time. gemma-3-4b-it writes \xThreecFourbcottokens\ tokens of chain of
thought for \xThreecFourbcotaccseven, at \xThreecFourbcotratio$\times$ the layer's GPU time. Program of thought, which
solves this benchmark with GPT-4~\cite{zhang2024nlep}, reaches only \xThreecFourbpotaccseven\ at 4B
(\xThreecFourbpotratio$\times$), and at 1B its programs fail (\xThreecOnebpotaccseven). The 4B models are a reference point, not a matched
baseline.

\textbf{What we learn.} The saving is the decoding that the layer avoids. Every generated token is a forward pass
that reads all of the model's weights, while the layer's own kernels add only \xThreeblayerextra\,ms to a plain
forward pass on arithmetic. The text tool's cost therefore grows with the length of its call, and the layer's does
not: from three to seven people its kernel time changes only from \xThreeclayerkernelthree\ to
\xThreeclayerkernel\,ms. Batching narrows the gap where the prompt dominates, and at batch size sixteen the text tool
uses \xThreectoolratiob$\times$ the layer's kernel time per item. These ratios hold for this implementation and range
of tapes. Longer tapes or more rounds would raise the layer's cost.

\begin{table*}[t]
\caption{\textbf{Cost at near-equal accuracy on the development task} (gemma-3-1b-it unless named; RTX 4090, bf16; four
to six operators). Wall-clock: the standard decoder with a dynamic cache. GPU kernels: summed kernel time from the
profiler. B1, B16: batch size one and sixteen. Floor: the model's weights read once per forward pass at 1\,TB/s.}
\label{tab:eff}
\small
\begin{tabular}{@{}lccccccc@{}}
\toprule
 & \multicolumn{2}{c}{accuracy} & \multicolumn{2}{c}{wall-clock (ms)} & \multicolumn{2}{c}{GPU kernels (ms)} & floor \\
per question & 4--6 ops & unseen templates & B1 & B16 & B1 & B16 & (ms) \\
\midrule
text tool: adapters copy the expression, exact evaluator \textit{(code)} & 0.99--1.00 & 0.93--0.98 & 235 & 255 & 103 & 138 & 32 \\
tape layer (frozen module after layer 8) & 1.00 & 0.55--0.59 & 20 & 25 & 8.9 & 16.0 & 2 \\
gemma-3-4b-it, chain of thought & 0.94 & -- & 3091 & 8010 & -- & -- & -- \\
\bottomrule
\end{tabular}
\end{table*}

\begin{table*}[t]
\caption{\textbf{Cost per item on the held-out task} (tracking shuffled objects, BIG-Bench Hard, 750 items; RTX 4090,
bf16). Accuracy on all items (chain of thought: Table~\ref{tab:state}); cost at seven people. Tokens: generated per
item (the tool's includes the answer step). Wall B1: Hugging Face \texttt{generate} with a dynamic cache, one item at a time, mean of
eight. Kernels B1: summed GPU kernel time from the profiler, mean of two. Wall B16: per item in a batch of sixteen.
Floor: forward passes $\times$ weight bytes at 1\,TB/s. $\times$\,layer: kernel time at batch 1 relative to the layer.
Program of thought: the model's program is run in a sandbox, with a call appended when it defines but never calls a
function (PAL's convention); the last printed line is matched to an option. \textit{(code)} in the tool: the
name-to-index map and the swaps.}
\label{tab:bbheff}
\small
\resizebox{\textwidth}{!}{\begin{tabular}{@{}lccccccccc@{}}
\toprule
 & \multicolumn{3}{c}{accuracy, people} & \multicolumn{6}{c}{per item at 7 people} \\
 & 3 & 5 & 7 & tokens & wall B1 (ms) & kernels B1 (ms) & wall B16 (ms) & floor (ms) & $\times$ layer \\
\midrule
tape layer: frozen swap module after layer 8, writer taught which tokens to mark & 1.00 & 1.00 & 1.00 & 0 & 31 & 11.0 & 5.1 & 2 & 1 \\
text tool: adapters write the names, code runs the swaps & 1.00 & 0.98 & 0.99 & 17 & 328 & 112.4 & 23.5 & 34 & 10 \\
gemma-3-1b-it, chain of thought & 0.85 & 0.57 & 0.34 & 485 & 8099 & 2842.8 & 597.3 & 970 & 257 \\
gemma-3-1b-it, program-of-thought & 0.05 & 0.07 & 0.05 & 439 & 7372 & 2124.9 & 838.5 & 878 & 192 \\
gemma-3-4b-it, chain of thought & 1.00 & 0.96 & 0.93 & 534 & 11832 & 6864.0 & 927.1 & 4596 & 621 \\
gemma-3-4b-it, program-of-thought & 0.92 & 0.80 & 0.76 & 252 & 5642 & 3807.4 & 471.1 & 2166 & 345 \\
\bottomrule
\end{tabular}}
\end{table*}

\subsection{RQ3: Learning the interface from answers}
\label{sec:writer}

Tape labels are rarely available. We therefore ask how the writer can be trained from the answer alone.

\textbf{Gradients through the module fail.} Trained end to end through the soft module, no tape is ever exact
(\eTwoetwoeexact) and the layer is no better than the adapters alone (\eTwoeTwoeFourSix). Worse, a writer trained
on labelled tapes loses its tape accuracy when updated using only answer supervision: by straight-through gradients (tapes exact
\eTwobwarmstexact), by two estimators intended to reduce bias, ReinMax and I-MLE (\pThreewarmreinmaxexact\ and
\pThreewarmimleexact), and even with the adapters below the writer frozen. Since lower-bias estimators fail too, the cause lies in the
gradient through the module rather than in one estimator.

\textbf{Reward on the module's outcome works, given a prior.} Treating the module as a black box --- sampling discrete
tapes and rewarding right answers --- completes a writer started with 100 labelled steps (\eTwobwarmrl, tapes exact).
From a random writer it stays at chance (\eTwobscratch): practically no random tape is right, so nothing is rewarded.
The hand-written prior replaces the labelled start. Alone, it lets the writer game the learned module, which tolerates
surplus symbols that exact code rejects (swap \xOnepriorswap): a learned module is a more lenient judge than code. With
the validity check and sixteen samples per question, \xOneckOneSixpassed\ of 3 seeds reach exact, swap-equivalent
tapes without a single tape label (Table~\ref{tab:etwo}, bottom rows).

\textbf{Adapters let the model answer around the tape.} On the held-out task the same reward, after 100 labelled steps,
loses the writer (\xTwocrlten\ at ten swaps): the adapters answer well without the tape, so the reward no longer
requires it. Removing the adapters --- gemma frozen as released, only writer and reader trained --- restores the
recipe. With the name prior and no tape or token label, \xSixpassed\ of \xSixruns\ runs learn to drop the story's
introduction by themselves and answer every benchmark item exactly (\xSixbbh, exact tapes; reader off \xSixreaderoff).
The remaining run games the module with a prefix of exchanges that cancel, right on every training story and wrong on
the benchmark.

\textbf{A larger host.} The recipe carries unchanged to gemma-3-4b-it (layer after decoder layer 10 of 34, the same
relative depth; nothing tuned on this model). In \pSixpassed\ of \pSixruns\ runs the writer learns from answers alone and
the layer answers every benchmark item exactly (\pSixbbh), with the reader off below chance (\pSixreaderoff). The two
exact runs come from the prior that marks each name with probability one half and keeps a memory. Both runs whose prior
marks every name failed. The layer costs \pSixoverhead$\times$ the larger model's plain forward pass and
\pSixcotratio$\times$ less than its chain of thought on the same items.

\textbf{What we learn.} Answers can replace tape labels, but not hand-written guidance. The prior supplies each
token's symbol, and what reward learns is which tokens to mark. Reward worked with the
toolkit of program induction: treat the module as a black box, start from a prior that makes a rewarded sample
likely, constrain the tape language so that a lenient module cannot be gamed, and sample enough. One lesson is
specific to a tool inside the model. When adapters can answer without the tape, answer-only reward stops requiring
the tape, and the writer is lost. Reliability remains the weakness. The success counts span several training configurations,
and validation accuracy did not detect the gamed run.

\begin{table*}[t]
\caption{\textbf{The held-out task: tracking shuffled objects (BIG-Bench Hard, 750 items), never trained on.} gemma
rows: zero-shot, greedy. Middle rows: gemma-3-1b-it with adapters trained on our own stories (three seeds for the first
and third rows, two for the others), with and without the frozen swap module as a layer; the answer is read over the
option letters in the same forward pass. Bottom rows: gemma-3-1b-it frozen, no adapters, only the writer and reader
trained --- with taught marks (two seeds), and from answers alone after a hand-written name prior (two variants of the prior, runs split
by whether their benchmark tapes are exact). Chance at 3 / 5 / 7 people: 0.33 / 0.20 / 0.14.}
\label{tab:state}
\small
\begin{tabular}{@{}lccccccc@{}}
\toprule
 & \multicolumn{3}{c}{BBH, people (= swaps)} & synthetic & tapes exact & reader off & tokens \\
model & 3 & 5 & 7 & 7 people, 10 swaps & BBH 7 & BBH 7 & generated \\
\midrule
gemma-3-1b-it, answers directly & 0.32 & 0.21 & 0.13 & -- & -- & -- & 4 \\
gemma-3-1b-it, chain of thought & 0.85 & 0.57 & 0.34 & -- & -- & -- & 492 \\
gemma-3-4b-it, answers directly & 0.31 & 0.12 & 0.08 & -- & -- & -- & 5 \\
gemma-3-4b-it, chain of thought & 1.00 & 0.96 & 0.93 & -- & -- & -- & 536 \\
\midrule
1B + adapters on our stories, no tape & 0.98--1.00 & 0.93--0.95 & 0.78--0.83 & 0.46--0.54 & -- & -- & 0 \\
\quad + frozen swap module as a layer, tape labels & 1.00 & 0.96--0.98 & 0.80--0.86 & 0.59--0.60 & 0.00 & 0.76--0.86 & 0 \\
\quad + the writer's marks taught too & 1.00 & 1.00 & 1.00 & 1.00 & 1.00 & 0.15--0.24 & 0 \\
\quad\quad marks for 100 steps, then answers only & 1.00 & 0.98--1.00 & 0.66--0.86 & 0.54--0.62 & 0.00 & 0.24--0.86 & 0 \\
\midrule
1B frozen, no adapters; marks taught & 1.00 & 1.00 & 1.00 & 1.00 & 1.00 & 0.14 & 0 \\
\quad from answers alone, 5 of 6 runs & 1.00 & 1.00 & 1.00 & 1.00 & 1.00 & 0.14 & 0 \\
\quad\quad the other 1: the module gamed & 0.00 & 0.20 & 0.14 & 1.00 & 0.00 & 0.14 & 0 \\
\bottomrule
\end{tabular}
\end{table*}

\subsection{RQ4: Transfer to rewritten prompts}
\label{sec:open}

The benchmark's wording is templated. A hand-written pattern parses all of it, so exactness on it says little about
reading language. We therefore had one language model (gpt-oss-120b) rewrite the first 100 items of each split under
fixed rules --- same people, objects, exchanges and question; no ``swap'', ``switch'' or ``trade'', no ``First'',
``Then'', ``Finally'' --- and a second model of another family (Mistral-Medium) check each rewrite against the original.
\xSevenkept\ of \xSevenrewrites\ rewrites agree, and the hand-written parser reads \xSevenparser\ of them. Nothing is
trained on them (Table~\ref{tab:open}).

The layers on the frozen model, without adapters, answer \xSevennoadapters\ of the rewrites --- \xSevenlayerfreeseven\ at
seven people when learned from answers alone, and \pSixopen\ in gemma-3-4b-it. On the same items, gemma-3-4b-it with
chain of thought falls from \xSevencotFourbsevenorig\ in the original wording to \xSevencotFourbseven, and the text tool
from \xSeventoolsevenorig\ to \xSeventoolseven. The writer trained together with adapters, exact on the original wording,
reaches only \xSevenlayerloraseven.

\textbf{What we learn.} Writers on an otherwise frozen host read unseen wording better than writers trained
together with adapters, and better than a four times larger model reasoning in text. A likely reason is that adapters
fit the training templates, although these runs do not isolate it. The rewrites keep the names and the structure of
the task, and only rewrites accepted by the judge are evaluated.

\begin{table*}[t]
\caption{\textbf{Open wording.} \xSevenkept\ judged rewrites of tracking shuffled objects (the first 100 items per
split; the hand-written parser reads none), and the same items in their original wording. Nothing is trained on them.
Layers read the answer in one forward pass (seeds or runs as in Table~\ref{tab:state}; ranges over them). ``Tape
right'': the written tape, run through exact code, gives the answer. ``Fires'': the share of rewrites on which the layer
acts (otherwise the base model answers). Chance at 3 / 5 / 7 people: 0.33 / 0.20 / 0.14.}
\label{tab:open}
\small
\resizebox{\textwidth}{!}{\begin{tabular}{@{}lcccccccc@{}}
\toprule
 & \multicolumn{3}{c}{rewritten, people} & \multicolumn{3}{c}{original, people} & tape right & fires \\
 & 3 & 5 & 7 & 3 & 5 & 7 & rewritten, 7 & rewritten \\
\midrule
layer, adapters, marks taught & 0.43--0.81 & 0.40--0.86 & 0.27--0.85 & 1.00 & 1.00 & 1.00 & 0.21--0.82 & 1.00 \\
layer, no adapters, marks taught & 0.94--0.96 & 0.94 & 0.87--0.88 & 1.00 & 1.00 & 1.00 & 0.87--0.88 & 1.00 \\
layer, no adapters, from answers alone & 0.94--0.95 & 0.91 & 0.90--0.94 & 1.00 & 1.00 & 1.00 & 0.90--0.94 & 1.00 \\
\quad with the prompt gate of Table~\ref{tab:silent} & 0.67 & 0.45 & 0.62--0.63 & 1.00 & 1.00 & 0.97--1.00 & 0.76--0.78 & 0.39--0.53 \\
\quad\quad gate trained with rewrites of our stories & 0.84--0.86 & 0.83--0.84 & 0.82--0.83 & 1.00 & 1.00 & 0.97--1.00 & 0.76--0.78 & 0.73--0.88 \\
text tool (adapters write the names, code swaps) & 0.65 & 0.64 & 0.68 & 1.00 & 1.00 & 0.99 & -- & -- \\
gemma-3-1b-it, chain of thought & 0.51 & 0.38 & 0.22 & 0.85 & 0.53 & 0.37 & -- & -- \\
gemma-3-4b-it, chain of thought & 0.75 & 0.79 & 0.71 & 0.99 & 0.94 & 0.96 & -- & -- \\
\bottomrule
\end{tabular}}
\end{table*}

\subsection{RQ5: Layers trained apart, composed}
\label{sec:two}

Modules trained apart are useful only if they can be put together. We place both frozen modules --- arithmetic and
state tracking --- as two layers after layer 8 of one gemma-3-1b-it, used as released. Each layer brings only its writer
and reader, is trained alone and never sees the other task.
At test time both writers run on every prompt, and neither is switched off by hand. A layer abstains by writing an
empty tape. Each writer is trained to mark nothing on general text, with prompts from the Pile~\cite{gao2020pile} as negatives. If a
writer marks no token, a hand-written rule sets $g=0$, the layer adds nothing, and the model is the base model, bit
for bit. Below we show that this alone lets the layers act too often, and add a check on the tape and a learned gate.

\textbf{Composition.} Each layer alone is as exact as before (\xFivearithFourSixalone\ at four to six operators,
\xFiveswapbbhsevenalone\ on the benchmark's seven-person items), and with both attached every split is answered exactly
as with its own layer alone (Table~\ref{tab:two}). A writer marks the other task's prompts at most 1\,\% of the time, and
with the reader off the answers fall to chance (\xFivereaderoffarith\ and \xFivereaderoffswap).

\textbf{Silence.} The cost of composition is false firing. The arithmetic writer fires on \xFivefiregeneralarith\ of
general text and the two writers on \xFivefiremmluarith\ and \xFivefiremmluswap\ of MMLU prompts, where the injected
result changes some answers: MMLU falls from \xFivemmlubase\ to \xFivemmlu. Two changes, one hand-written and one learned, remove
this (Table~\ref{tab:silent}). Requiring the tape to compute something --- an operator, an exchange --- removes most
firing on look-alike prompts, which was mostly lone numbers and lone names. A learned gate on the whole prompt does the
rest: one logit from the mean state after layer 8, trained on the layer's own prompts against general text, questions
and look-alikes from training sources only. With it neither layer fires on any of \xFivecprompts\ prompts --- the Pile,
BoolQ~\cite{clark2019boolq}, two samples of MMLU, logical deduction and web of lies from BBH, and the math word
problems of GSM8K~\cite{cobbe2021gsm8k} and SVAMP~\cite{patel2021svamp} --- MMLU is exactly the base model's
(\xFivecgatemmlu), and both tasks are unchanged.

\textbf{What we learn.} Layers trained apart can share one frozen model without disturbing it or each other, as
long as each knows when not to act. That takes a tape language that demands a computation and a gate that reads the
whole prompt. The two layers select their tasks independently and do not cooperate on one computation. Conservative
activation has a cost on open wording. The gated layer acts on \xSevencgbothfire\ of the rewrites and reaches
\xSevencgbothacc\ overall, against \xSevencwherefired\ where it acts. Both the prompt gate and the writer's tape
checks can reject a valid prompt, and the tape checks remain the larger source of misses (Appendix~\ref{app:more}).
The best open-wording results of \S\ref{sec:open} come from a different, ungated configuration.

\begin{table*}[t]
\caption{\textbf{Two frozen modules as two layers of one frozen gemma-3-1b-it}, each layer trained alone, no adapters
(two seeds, used hard). ``Together'': both layers attached. ``Fires'': the share of prompts on which a writer marks a
token --- in the task rows the other task's writer, in the bottom rows the arithmetic / the swap writer.
\textit{Hand-written:} the packing, the gate (an empty tape adds nothing), the tape languages.}
\label{tab:two}
\small
\begin{tabular}{@{}lccccc@{}}
\toprule
 & alone & together & change & tapes exact & fires: other layer; arithmetic / swap writer \\
\midrule
arithmetic, 4--6 operators & 1.00 & 1.00 & +0.000 & 1.00 & 0.00 \\
\quad unseen templates & 0.39--0.73 & 0.39--0.73 & +0.000 & 0.34--0.69 & 0.01 \\
\quad 12 operators & 0.96--1.00 & 0.96--1.00 & +0.000 & 0.95--0.99 & 0.00 \\
state tracking, 7 people, 10 swaps & 1.00 & 1.00 & +0.000 & 1.00 & 0.00 \\
\quad BBH, 3 people & 1.00 & 1.00 & +0.000 & 1.00 & 0.00 \\
\quad BBH, 5 people & 1.00 & 1.00 & +0.000 & 1.00 & 0.00 \\
\quad BBH, 7 people & 1.00 & 1.00 & +0.000 & 1.00 & 0.00 \\
\midrule
MMLU (base model 0.382) & -- & 0.368--0.370 & $-0.014$ to $-0.012$ & -- & 0.05--0.09 / 0.04--0.06 \\
general text (the Pile, 10k snippets) & -- & -- & -- & -- & 0.07--0.09 / 0.00 \\
\bottomrule
\end{tabular}
\end{table*}

\begin{table*}[t]
\caption{\textbf{Keeping two layers silent on general text} (both layers attached, two seeds each). ``Fires'': the share
of prompts on which the more active writer adds its result --- general text: the Pile and BoolQ; BBH logic: logical
deduction with five objects (no exchanges); GSM8K: word problems (test split). None of these is trained on, except that
the gate's negatives include GSM8K's training split. \textit{Hand-written:} the well-formedness check and the
requirement that the tape computes; \textit{learned:} the question negatives and the gate.}
\label{tab:silent}
\small
\resizebox{\textwidth}{!}{\begin{tabular}{@{}lccccc@{}}
\toprule
 & MMLU (base 0.382) & fires: general text & fires: BBH logic & fires: GSM8K & BBH tracking, 7 \\
\midrule
writer trained on general text; marks nothing $\Rightarrow$ adds nothing & 0.368--0.370 & 7.2--9.0\,\% & 86.8--92.4\,\% & 54.0--80.8\,\% & 1.00 \\
\quad + question negatives, well-formed tape & 0.376--0.386 & 7.2--9.0\,\% & 36.0--38.0\,\% & 28.2--37.2\,\% & 0.99--1.00 \\
\quad + the tape must compute (an operator, a swap) & 0.382 & 0.4\,\% & 0.0--4.8\,\% & 0.2--0.6\,\% & 0.99--1.00 \\
\quad + a learned gate on the whole prompt & 0.382 & 0.0\,\% & 0.0\,\% & 0.0\,\% & 0.99--1.00 \\
\bottomrule
\end{tabular}}
\end{table*}

\section{Discussion and limitations}

The strongest result is that a bounded computation can be accessed through hidden states at substantially lower
measured cost than generating its inputs as text. The learned executor is accurate enough for the evaluated tasks.
The harder problem is constructing its input reliably. This distinction explains why exact benchmark performance,
failed training runs and missed calls on new wording can coexist.

The evidence has three separate scopes. The layer with a writer trained on labelled tapes supports the main cost comparison. The answer-only
experiments establish that tape annotations can sometimes be replaced by priors and reward. The gated composition
experiments establish preservation on tested negative prompts, with lower coverage on open wording. No single
configuration has yet demonstrated all three benefits reliably across new tasks.

\textbf{When a learned module, when code.} On both tasks the learned module is interchangeable with exact code, so
code would have served. A learned module earns its place where no code exists, or where modules should be trained and
exchanged independently. Our experiments show that independently trained modules can coexist (\S\ref{sec:two}). They do not show that
learning the executor is preferable to implementing it in code.

\textbf{Limitations.}
\begin{itemize}[leftmargin=*,noitemsep]
\item One model family: gemma-3-1b-it, and gemma-3-4b-it for the label-free state-tracking layer only (exact in
  \pSixpassed\ of \pSixruns\ runs there); two seeds for most layer configurations.
\item Both tasks have known exact algorithms. Object tracking was selected after the initial design, but informed
  later interface decisions. A new task evaluated only after freezing the final design is still needed for a broader
  transfer claim.
\item The open-wording rewrites are written by one language model and judged by another. They keep the names, so
  coreference is not tested.
\item The layer fires once, at the last prompt position, and returns one symbol. Calls in the middle of a generated
  solution and results of several tokens are untested. This is the step to tasks such as math word problems.
\item The prompts contain few distractors, numbers or names that are not part of the computation. On the one held-out
  arithmetic template with an unrelated number, the writer copies it and fails (Appendix~\ref{app:more}).
\item Training the writer from answers is not yet reliable, and validation accuracy did not detect a gamed tape.
\item Exchanging the module for exact code shows equivalence on the evaluated inputs, not on every possible tape. That no
  layer fired on the negative prompts does not guarantee silence on arbitrary text.
\item The text tool was decoded token by token. Speculative decoding~\cite{leviathan2023speculative}, which verifies
  several drafted tokens per forward pass, would reduce its cost and was not measured.
\item The swap and reader-off checks do not edit the tape. Changing the written tape and checking that the answer
  follows would test reliance on the module more directly.
\item Latency was measured on one GPU, with small timing samples and batch sizes one and sixteen, not in a serving
  system. The ratios exclude training cost and do not establish a hardware-independent speedup.
\end{itemize}

\section{Conclusion}

A separately trained tape module can supply a language model with the result of a bounded computation during one
forward pass. On object tracking, with a writer trained on labelled tapes, the layer achieves near-equal accuracy to a text-tool baseline at about
$1/\xThreectoolratio$ of its GPU kernel time at batch size one. Exchanging the module for exact code shows that it computes correctly, and removing its result shows that the
answer depends on it. Answer-only training and gated composition extend this result, but expose
the remaining difficulty: reliably extracting the right tape and activating it on unfamiliar wording. The immediate
research priority is to test a fixed interface on fresh tasks and wording. Calls during generation and multi-symbol
outputs would extend the scope beyond the computation studied here.

\bibliographystyle{ACM-Reference-Format}
\bibliography{refs}

\appendix

\section{Further results}
\label{app:more}

\textbf{The module alone.} Without the length curriculum \eOnebmixsoftformed\ of 4 seeds form. Trained straight-through,
the selector never leaves the first stage (\eOnebcurstsix), because hard no-op rounds give the choice no gradient. The
reverse holds for a single column adding numbers: there straight-through forms (\eOnestformed\ of 4) and soft training
does not survive use (\eOnesoftfour). Which relaxation a module needs is therefore an empirical question. Surplus rounds
become no-ops, and twice the rounds at use change nothing (\eOnebcursoftTwoxtwentyfour).

\textbf{Unseen templates on the development task.} On two templates never trained on the writer reaches
\eTwosupFourSixheld. Per template, it is exact on one (\eTwosupheldplease) and fails the other (\eTwosupheldafter, tapes
exact \eTwosupheldafterexact), ``After $d$ hours the result of \dots'', where it also copies the number of hours.
Trained on twelve templates instead of four, it reaches \xFourtOneTwokEightheld\ on eight held-out templates, against
\xFourtFourkEightheld\ with four. A writer reading layer 13 instead of 8 does worse (in range \xFourtFourkOneThreerange,
tapes exact \xFourkthirteenexact). The text tool, copying through the model's own generation, generalises to unseen
templates (\xThreetoolheld) where a writer trained on four templates does not.

\textbf{Gradient estimators.} A small tape loss kept beside the answer does not save a trained writer
(\eTwobwarmauxexact), nor does freezing the adapters below it (\xOnefrozenlow). With those adapters frozen, ReinMax and
I-MLE with its perturbation scale adapted per step lose a trained writer (accuracy \pThreewarmall) and neither learns
from scratch (\pThreescratch). For I-MLE only the symbols received a usable gradient, and the scale needed for the marks
diverged.

\textbf{Label-free training on the development task.} The prior followed by reward reaches \xOneprior, but with tapes
exact only \xOnepriorexact. The validity check lets \xOnebpassed\ of \xOnebruns\ runs find the exact tape, and the rest stay
at \xOnebstuck. With sixteen samples the passing seeds hold at twelve operators (\xOnectwelve). A memory of rewarded
tapes manages \xOnecmempassed\ of 3.

\textbf{Label-free training on the held-out task.} The prior marks every token that is one of the seven names and
writes its index. It marks the introduction and the assignments as well, twice as many names as the tape holds. Reward
is given only for a well-formed tape whose marks fit the cells. The exact runs reach validation accuracy above 0.95 by
step \xSixstep. The gaming run writes, before the true tape, a prefix of exchanges that cancel (one name paired with
itself, then one pair twice): right on every training story (\xSixgamesyn) and therefore chosen by validation accuracy,
wrong on the benchmark, where the same prefix breaks the pairing (\xSixgamebbh). Preferring the shortest rewarded tape,
to rule such prefixes out, fails in all six runs: the shortest rewarded tape is usually one that is right by chance ---
the asked name alone --- and the writer learns to mark almost nothing. In gemma-3-4b-it one exact run found its tape
only in the last 500 of 2500 steps. Of the failed runs, one collapsed to a one-symbol tape, the other kept names of the
introduction before the true tape (BBH \pSixfailbbh). There the layer costs \pSixkernel\ against \pSixplain\,ms of GPU
kernel time at seven people, batch 1, fires on no general text and on at most \pSixmmlufire\ of MMLU, where accuracy is
\pSixmmlu\ against \pSixbasemmlu\ for the base model.

\textbf{Silence, step by step.} Half the negatives replaced by questions (the MMLU auxiliary training set, disjoint from
its test) silence the swap writer on MMLU (\xFivebqneganyswapmmlu) but not the arithmetic writer
(\xFivebqneganyarithmmlu), and a check that the tape is well formed barely helps: \xFiveblone\ of the arithmetic firings
it passes on MMLU are a single digit, which is a well-formed expression. On look-alike prompts the swap writer fires on
\xFivebqnegvalidlogic\ of the logical-deduction puzzles of BIG-Bench Hard (people and objects, no exchanges), the
arithmetic writer on \xFivebqnegvalidgsm\ of GSM8K word problems. Requiring the tape to compute brings this to
\xFivecstrictlogic\ and \xFivecstrictgsm. The gate's look-alike negatives are GSM8K's training split and puzzles of our
own with names and no exchanges. MMLU equals the base model on a second sample as well (\xFivecgatemmlufresh), and the tasks
keep \xFivecbbh\ on the benchmark. Without adapters the arithmetic writer is weaker on unseen wording
(\xFivearithFourSixheldalone).

\textbf{The composable layer on open wording.} Trained on our templates only, its gate passes \xSevenlayergatedfire\ of
the rewrites, and where it passes the layer answers \xSevenlayergatedwherefired. Retrained with \xSevenbpos\ rewrites of
our own stories as further positives and \xSevenbneg\ rewritten name puzzles as further negatives, it passes
\xSevenbgbothfire\ and the layer answers \xSevenbgbothacc, still silent on general and look-alike prompts. A threshold
set on \xSevenccalsets\ held-out negative sets rises rather than falls: three questions of BIG-Bench Hard's
colored-objects task score like task prompts, while every other calibration set stays below \xSevencothermax. The gate
alone passes \xSevencgatepass\ of the rewrites. On part of them the writer's tape fails the tape checks, so the layer
stays silent and the base model answers. Open wording needs a writer trained on more varied wording more than a better
gate.

\clearpage
\onecolumn
\section{Registered predictions and outcomes}
\label{app:pred}

Each prediction was registered in the experiment's configuration before the run. A label names that configuration
in the code repository and the prediction's letter within it (\texttt{x5-c}: prediction c of experiment X5). The
parenthesis says what was measured. Rows appear in the order the experiments ran. ``Arm'': one configuration of an
experiment. ``Warm'': a writer first trained on labelled tapes.

\small
\begin{longtable}{@{}p{0.21\textwidth}p{0.41\textwidth}p{0.29\textwidth}@{}}
\caption{\textbf{Registered predictions and their outcomes.}}
\label{tab:pred}\\
\toprule
label (measurement) & prediction & outcome \\
\midrule
\endfirsthead
\toprule
label (measurement) & prediction & outcome \\
\midrule
\endhead
e2e1b-a (operator order) & $\ge 0.95$ at 6 operators on 3 of 4; $\ge 0.90$ at 24 on 2 of 4 & confirmed (4 of 4; \eOnebcursofttwentyfour) \\
e2e1b-b, c (curriculum; straight-through) & without curriculum $\ge 0.20$ lower; straight-through $\le 0.30$ & refuted on the median (forms on 2 of 4); confirmed \\
e2e1b-d, e (no-ops) & twice the rounds change $\le 2\,\%$; cost $\le 0.05$ & confirmed \\
e2e2-a, b (layer) & adapters alone $\le 0.50$; tape labels $\ge 0.95$, swap within 0.02 & confirmed (\eTwodirectFourSix; \eTwosupFourSix, swap identical) \\
e2e2-c (answers only) & $\ge 0.80$ on one seed & refuted (\eTwoeTwoeFourSix, no exact tape) \\
e2e2-d, e (warm start; reader off) & warm within 0.05 of labels; reader off $\le 0.30$ & confirmed, d only through the checkpoint kept while supervised \\
e2e2-f (beyond range) & every tape arm $\le 0.50$ at 8 and 12 operators & refuted: labelled writers hold (1.00) \\
e2e2b-a (from scratch) & neither straight-through nor reward learns & confirmed (\eTwobscratch) \\
e2e2b-b, c (reward keeps; completes) & after 600 or 100 labelled steps, reward keeps or completes the writer & confirmed (\eTwobwarmrl, tapes exact) \\
e2e2b-d, e (straight-through; small tape loss) & straight-through unlearns it; a 0.1 tape loss keeps it & confirmed; refuted (\eTwobwarmauxexact) \\
e2e2b-f (generalisation) & reward costs nothing on unseen templates & confirmed \\
x1-a, b (copy prior) & a rewarded sample likely at start; prior + reward $\ge 0.95$ with exact tapes; maximum marginal likelihood $\ge$ REINFORCE & a refuted on tapes (\xOnepriorexact: the writer games the module); b confirmed \\
x1-c, m (labelled steps; frozen adapters) & 30 labelled steps suffice, 10 do not; freezing lower adapters does not save the writer & refuted (both stay at chance); confirmed \\
x1b-a (validity check) & exact, swap-equivalent tapes on 1 of 2 seeds per arm & refuted as registered (\xOnebpassed\ of \xOnebruns\ runs) \\
x1b-b, c, d (validity check, continued) & exact on the hard templates; 12 operators; unseen templates $\ge 0.55$ where (a) holds & confirmed \\
x1c-a, b (exploration) & 16 samples on 2 of 3 seeds; memory on 2 of 3 & confirmed (\xOneckOneSixpassed\ and \xOnecmempassed\ of 3) \\
x1c-c, d (both aids together) & both together on 3 of 3; 12 operators $\ge 0.95$ where they hold & refuted (\xOneckOneSixmempassed\ of 3); confirmed \\
x4-a, b (wording) & 12 templates $+0.20$ on unseen ones; a later layer $+0.10$ & refuted by the margin (reaches \xFourtOneTwokEightheld); refuted \\
x4-c, d (wording, continued) & best arm $\ge 0.90$ unseen; every arm $\ge 0.99$ in range & confirmed; refuted (layer 13: \xFourtFourkOneThreerange) \\
x3-a, c, d (efficiency) & text tool $\ge 0.95$; 4B chain of thought $< 0.95$ and $\ge 4\times$ slower; tool $\ge 0.90$ at 12 & confirmed \\
x3-b, x3b-a (efficiency) & layer $\ge 2\times$ faster, also with an optimised decoder (read on kernel time) & confirmed (\xThreeratio$\times$; \xThreebkernelratio$\times$) \\
x3b-b (optimised decoder) & optimising the decoder helps the text tool more than the layer & refuted \\
x2a-a, b, d (room for a tool) & 1B near chance at 7 people, chain of thought $\ge 0.20$ below the hand-written parser; the parser reads all items & confirmed \\
x2a-c (4B reasoning) & 4B with chain of thought $\le 0.70$ at 7 people & refuted (\xTwoafourcot) \\
x2b-m-a, b (swap module) & exact at 6 swaps on 3 of 4 seeds; $\ge 0.95$ at 14 on 2 of 4 & refuted (\xTwobmformed\ of 4); confirmed \\
x2b-l-a (adapters alone, no tape) & $\le 0.40$ at 7 people & refuted (\xTwobdirectbbhseven) \\
x2b-l-b--e (the layer) & tape used and transferring to the benchmark & not meaningful (tape unused, \xTwobsupexact\ exact); e refuted \\
x2c-a, b, c (marks taught) & tapes exact; $\ge 0.85$ at 10 swaps; benchmark tapes $\ge 0.80$ and above adapters & confirmed (\xTwocall) \\
x2c-d (answers only after 100 steps) & within 0.05 of the labelled writer at 10 swaps & refuted (\xTwocrlten) \\
x3c-a, b (benchmark cost) & layer $\ge 0.99$; $\ge 100\times$ faster than 4B chain of thought at batch 1, $\ge 20\times$ at 16 & confirmed (\xThreecFourbcotratio$\times$; \xThreecFourbcotratiob$\times$) \\
x3c-c, d (program-of-thought) & 4B $\ge 0.90$ at 7 people, 1B $\le 0.80$; 4B $\ge 100$ tokens and $\ge 50\times$ the layer & refuted (\xThreecFourbpotaccseven); confirmed \\
x3c-e, f (text tool; growth) & tool $\ge 0.95$, layer $\ge 5\times$ faster; generating arms grow with swaps, the layer within $1.2\times$ & confirmed (\xThreectoolratio$\times$) \\
p3-a, b, c (estimators; drift excluded) & ReinMax and I-MLE lose a trained writer; neither learns from scratch; straight-through with the lower adapters frozen loses it too & confirmed (tapes exact 0.00; for I-MLE only the symbols got a usable gradient) \\
p2-a, b (third seed, arithmetic) & tape labels $\ge 0.95$, exact, swap-equivalent, reader off $\le 0.30$; adapters alone $\le 0.50$ & confirmed \\
x5-a, b, e (two layers, no adapters) & each alone at its earlier level; together within 0.01; reader off $\le 0.30$ & confirmed (identical on every split) \\
x5-c, d (firing; general ability) & other task $\le 1\,\%$; MMLU within 0.01 of base & confirmed; refuted (\xFivemmlu\ against \xFivemmlubase) \\
p2-c, d (third seed, benchmark) & marks taught: exact, $\ge 0.99$ on every subset; adapters alone 0.70--0.90 at 7 people & confirmed \\
x5b-a (well-formed tape only) & MMLU firing $\le 3\,\%$ per writer, drop at most half of that without the check & refuted (\xFivebxfivevalidarithmmlu, \xFivebxfivevalidswapmmlu) \\
x5b-b, c (question negatives) & $\le 1\,\%$ on MMLU, MMLU within 0.005 on both seeds; near negatives $\le 2$--$5\,\%$ & refuted (arithmetic \xFivebqneganyarithmmlu; MMLU \xFivebmmlu; logic \xFivebqnegvalidlogic) \\
x5b-d (composition) & together = alone, cross-firing $\le 1\,\%$ & confirmed \\
x5c-a (the tape must compute) & MMLU firing $\le 1.5\,\%$, MMLU within 0.005; near negatives barely move & confirmed, except the last: refuted (logic \xFivecstrictlogic, GSM8K \xFivecstrictgsm) \\
x5c-b, c, d (prompt gate) & near negatives $\le 2\,\%$, general $\le 1\,\%$, MMLU within 0.005; tasks kept; together = alone & confirmed (no firing; MMLU identical) \\
x6-a (answers only, prior near 1) & below 0.60 at 5 people on both seeds & refuted (1.00 on both) \\
x6-b, c, d (answers only, explore) & exact on 1 of 2 seeds; BBH 7 $\ge 0.80$ with exact tapes; reader off at chance & confirmed (2 of 2; \xSixbbh; \xSixreaderoff) \\
x6b-a, c (shortest rewarded tape; a third seed) & every run with the preference exact on BBH; plain seed 2 exact on 1 of 2 & refuted (0 of 6, at chance); confirmed (2 of 2) \\
x7-a (rewrites) & $\ge 70\,\%$ kept; the parser reads $\le 20\,\%$ & confirmed (\xSevenkept\ of \xSevenrewrites; \xSevenparser) \\
x7-b (adapter layer on rewrites) & $\ge 0.80$ at 3 people, $\ge 0.60$ at 7, on one seed & confirmed (\xSevenlayerlorathree; \xSevenlayerloraseven) \\
x7-c (who loses more) & layers lose $\ge 0.15$ at 7, chain of thought at 4B $\le 0.05$ & refuted (4B loses \xSevencotFourbdrop; layers without adapters \xSevenlayerfreedrop--\xSevenlayersupdrop) \\
x7-d, e (answers alone; tool) & answers-only within 0.10 of taught marks; tool $\ge$ the adapter layer at 7 & confirmed; holds for 2 of 3 seeds \\
x7-f (prompt gate) & passes $\ge 80\,\%$ of rewrites; where it passes, within 0.05 of taught marks & refuted (\xSevenlayergatedfire); confirmed (\xSevenlayergatedwherefired) \\
x7b-a, b (gate with rewritten stories) & passes $\ge 90\,\%$ of rewrites, answers $\ge 0.85$; look-alikes $\le 2\,\%$, MMLU within 0.005 & refuted (\xSevenbgbothfire; \xSevenbgbothacc); confirmed (no firing; MMLU identical) \\
x7b-c, d (the rewriter's style; tasks) & without rewritten negatives the style leaks onto look-alikes; tasks unchanged & refuted (no firing either way); confirmed \\
x7c-a, b (the gate's threshold) & a threshold set on held-out negatives falls below 0; then $\ge 90\,\%$ passed, answers $\ge 0.85$ & refuted (rises to \xSevencgbotht; \xSevencgbothfire; \xSevencgbothacc) \\
x7c-c, d (silence; tasks) & look-alikes $\le 2\,\%$, MMLU within 0.005; tasks unchanged & confirmed (\xSevencgbothnear; MMLU identical) \\
p6-a, b (4B host) & $\ge 2$ of 4 runs exact on BBH; reader off at chance & confirmed (\pSixpassed\ of \pSixruns; \pSixreaderoff) \\
p6-c, d, e (4B: open wording; cost; silence) & $\ge 0.85$ at 7 on rewrites; $\le 1/100$ of 4B chain of thought, $\le 1.3\times$ the plain forward; MMLU firing $\le 5\,\%$ & confirmed (\pSixopenseven; \pSixcotwallratio$\times$ wall, \pSixoverhead$\times$; \pSixmmlufire) \\
\bottomrule
\end{longtable}

\end{document}